\documentclass[10pt,twocolumn,letterpaper]{article}

\usepackage{cvpr}              
\usepackage{makecell}
\usepackage{multirow}

\definecolor{cvprblue}{rgb}{0.21,0.49,0.74}
\usepackage[pagebackref,breaklinks,colorlinks,allcolors=cvprblue]{hyperref}

\def\paperID{*****} 
\def\confName{CVPR}
\def\confYear{2025}

\title{S3D: Sketch-Driven 3D Model Generation}

\author{
Hail Song\textsuperscript{1*}\,
Wonsik Shin\textsuperscript{2*}\,
Naeun Lee\textsuperscript{2}\,
Soomin Chung\textsuperscript{2}\,
Nojun Kwak\textsuperscript{2\textdagger}\,
Woontack Woo\textsuperscript{1\textdagger} \\
\textsuperscript{1}KAIST \quad
\textsuperscript{2}Seoul National University \\
{\tt\small
\{hail96, wwoo\}@kaist.ac.kr\,
\{wonsikshin, better\_\_62, soomin200, nojunk\}@snu.ac.kr\,
}
}

\begin{document}
\twocolumn[{
    \maketitle
    \begin{center}
        \includegraphics[width=1\textwidth]{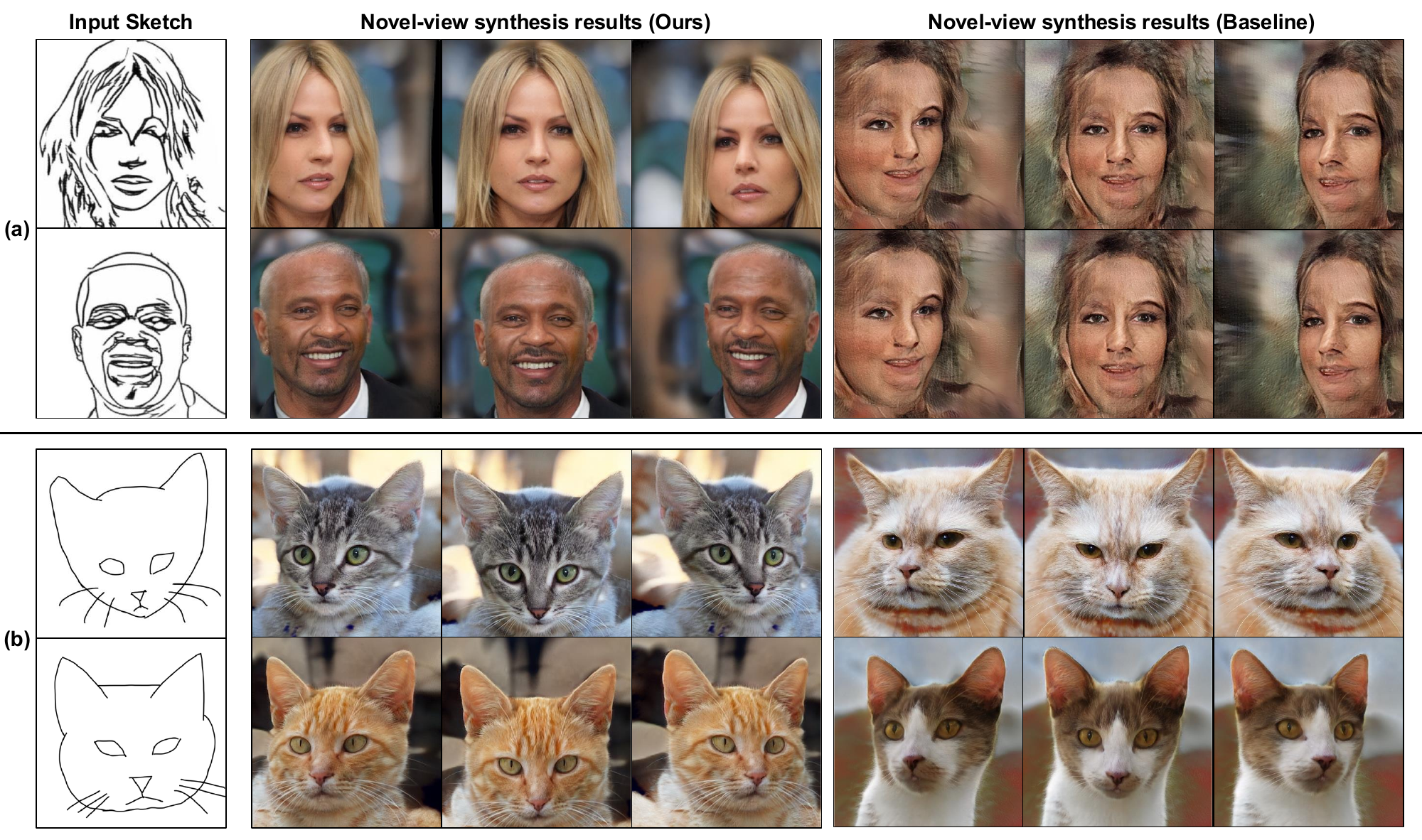}
        \captionof{figure}{\textbf{Comparison between our model and the baseline method.} (a) Given a single sketch image as input, \textbf{S3D} (Ours) generates a 3D human face that can be rendered from novel viewpoints. In contrast, the baseline method fails to reconstruct plausible 3D geometry, instead producing nearly identical structures from distinct human sketches.
        (b) While the baseline often produces models with misaligned facial outlines and unnatural neck contours, S3D generates 3D cat models with high-fidelity textures and more coherent shapes overall.
        }
        \label{fig:teaser}
    \end{center}
}]

\renewcommand{\thefootnote}{*}
\footnotetext{Equal contribution.}
\renewcommand{\thefootnote}{\textdagger}
\footnotetext{Corresponding author.}
\renewcommand{\thefootnote}{\arabic{footnote}}

\begin{abstract}
Generating high-quality 3D models from 2D sketches is a challenging task due to the inherent ambiguity and sparsity of sketch data. In this paper, we present \textbf{S3D}, a novel framework that converts simple hand-drawn sketches into detailed 3D models.
Our method utilizes a U-Net-based encoder-decoder architecture to convert sketches into face segmentation masks, which are then used to generate a 3D representation that can be rendered from novel views. To ensure robust consistency between the sketch domain and the 3D output, we introduce a novel style-alignment loss that aligns the U-Net bottleneck features with the initial encoder outputs of the 3D generation module, significantly enhancing reconstruction fidelity. 
To further enhance the network’s robustness, we apply augmentation techniques to the sketch dataset. 
This streamlined framework demonstrates the effectiveness of S3D in generating high-quality 3D models from sketch inputs. The source code for this project is publicly available at \textit{\textcolor{blue}{\href{https://github.com/hailsong/S3D}{https://github.com/hailsong/S3D}}}.
\end{abstract}
\vspace{-10pt}    
\section{Introduction}
\label{sec:intro}

The increasing demand for 3D content generation in industries such as film, gaming, and virtual reality has accelerated the development of neural rendering techniques. Neural rendering has been extensively utilized not only for novel view synthesis but also for 3D model generation from visual data.
3D model generation using such techniques often requires multi-view images, which can be impractical in many real-world scenarios. This limitation has led to studies that aim to reconstruct 3D models from a single image input~\cite{Chan2022eg3d}.
However, in many practical cases of 3D content creation, concrete image references are simply unavailable, whereas only high level conceptual descriptions exist.
As a result, research has shifted toward utilizing more abstract visual inputs—such as 2D segmentation maps—for 3D model generation~\cite{pix2pix3d}.
Despite these advancements, generating accurate 3D representations from highly abstract inputs—such as hand-drawn sketches—remains a significant challenge.

To address these challenges, we propose \textbf{S3D}, a novel end-to-end pipeline for \textbf{S}ketch-to-\textbf{3D} model generation. 
Leveraging a divide‐and‐conquer strategy, our method utilizes a U-Net-based encoder-decoder architecture to convert 2D sketch inputs into segmentation masks, which are then processed by a mask-to-3D module to generate high-fidelity 3D facial models from a single segmentation mask.
To enhance the robustness and accuracy of the reconstruction, we employ a combination of Cross-entropy loss and Dice loss. Furthermore, leveraging the prior knowledge embedded in pre-trained models, we introduce a novel loss function that aligns the style vector from the U-Net bottleneck with the initial encoder output of the mask-to-3D module.


Unlike previous approaches, which often struggled with sketch-based 3D face generation due to the complexity of human facial structures and the limited geometric cues in sketches, our model demonstrates the ability to generate high-fidelity 3D representations of both human and cat faces from simple sketch inputs. By enabling accurate 3D face reconstruction from such minimal input, it dramatically reduces the cost and time required for traditional 3D modeling, while unlocking innovative use cases such as efficient forensic facial montages and personalized avatar creation.

In summary, our key contributions are as follows:

\begin{itemize}
    \item We propose S3D, an end-to-end pipeline for sketch-based 3D face model generation. We integrate a U-Net and a tri-plane-based 3D model generation network to achieve high-quality outputs.
    \item We introduce a novel style-alignment loss and augmentation strategies to enhance the consistency between sketch-based segmentation masks and 3D reconstructions.
    \item We enable a new task of generating 3D human faces directly from facial sketches, which was previously unachievable.
\end{itemize}

\section{Related Work}
\label{sec:relatedwork}

\subsection{3D Neural Representation}
Neural rendering techniques, such as Neural Radiance Fields (NeRF)~\cite{mildenhall2021nerf}, leverage neural networks to represent and reconstruct 3D scenes continuously. 
NeRF learns a 3D scene representation by training on hundreds of images captured from varying camera transformations, enabling the generation of high-quality 3D renderings and facilitating tasks like novel view synthesis.
Recently, advancements in few-shot NeRF~~\cite{seo2023flipnerfflippedreflectionrays, yang2023freenerfimprovingfewshotneural} techniques have emerged, aiming to achieve comparable performance while  reducing the number of input images required for training. 

However, despite significant advances in neural 3D reconstruction strategies, these methods still require multiple images captured from different viewpoints under complex setup conditions.
This can limit the ease of converting real-world objects or scenes into 3D representations.

\subsection{3D Generation Using Neural Rendering}
Unlike methods that construct 3D models or avatars directly from visual information~\cite{grassal2022neural, zhao2024psavatar, rcsmpl, tran2024voodoo3d, saito2020pifuhd, song2024toward, alldieck2018video, an2023panohead, li2023generalizablegoha}, generative approaches operate in a latent space, where they must infer a diverse range of plausible outcomes—posing a fundamentally different challenge.
For basic 3D asset generation, recent studies have focused on inferring 3D representations from a single image using various image-to-3D translation methods.
Many of these approaches integrate various 3D representations directly into the learning pipeline, employing techniques such as voxel grids~\cite{henzler2021escapingplatoscave3d}, voxelized 3D features~\cite{nguyenphuoc2019hologanunsupervisedlearning3d}, and 3D morphable models~\cite{tewari2020stylerigriggingstylegan3d}. Other approaches such as StyleNerf~\cite{gu2021stylenerfstylebased3dawaregenerator} leverage NeRF-based models to generate 3D scenes.

Similarly, EG3D~\cite{Chan2022eg3d} introduces a semantic style vector, enabling the generation of high-quality 3D representations using tri-plane-based feature map.
Since then, the tri-plane representation has gained widespread adoption and has been widely applied and further developed in numerous 3D generation studies 
\cite{an2023panohead, chen2023mimic3d, gao2022get3d, wang2023rodin, yi2023progressive}.
Among recent approaches, pix2pix3D~\cite{pix2pix3d} and SketchFaceNerf~\cite{SketchFaceNeRF2023} utilize sketches or segmentation masks for 3D modeling. However, due to the abstract and ambiguous nature of sketches, these methods exhibit limited generation performance, particularly when dealing with complex structures such as human faces.



\begin{figure*} [ht]
  \centering
  \includegraphics[width=0.95\textwidth]{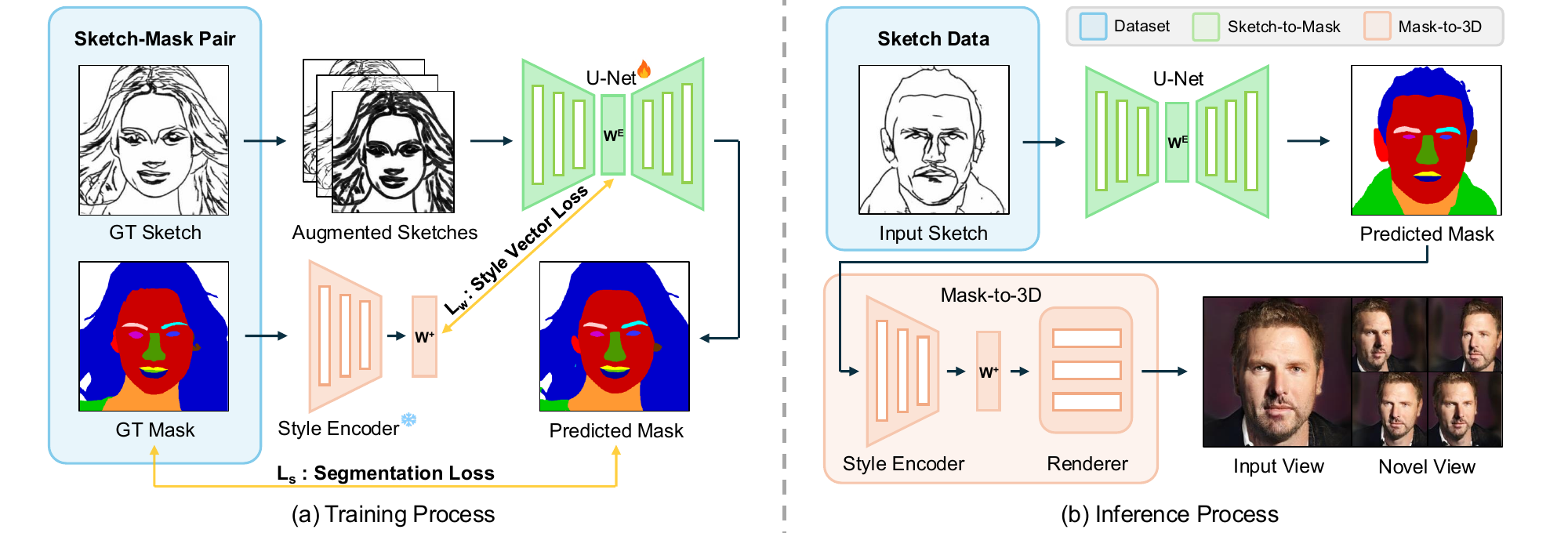}
  \caption{
  \textbf{Pipeline overview of S3D.} Our method combines a U-Net-based sketch-to-mask module with mask-to-3D module to generate 3D models. \textbf{(a) Training:} The U-Net is trained with Style Vector loss and Segmentation Loss to align latent features and improve mask accuracy. \textbf{(b) Inference:} A sketch is translated into a mask, which is then used by mask-to-3D module to synthesize 3D models.
}

  \label{fig:overview}
\end{figure*}
\section{Methods}
\label{sec:methods}

We propose \textbf{S3D}, a method for \textbf{S}ketch-to-\textbf{3D} generation by integrating a U-Net-based~\cite{ronneberger2015unetconvolutionalnetworksbiomedicalUNET} sketch-to-mask module with tri-plane-based 3D model generation module.

Our method converts sketches into segmentation masks via a sketch-to-mask module, which are then passed to a mask-to-3D module to generate 3D faces from a single segmentation mask. The following sections describe the sketch-to-mask module, the mask-to-3D module, and the associated training losses. \autoref{fig:overview} illustrates the full pipeline of S3D.

\subsection{Sketch-to-Mask Module}

The sketch-to-mask module adopts a CNN-based U‑Net architecture that directly maps 512 × 512 input sketches to pixel‑wise segmentation masks. Specifically, it comprises seven encoder-decoder pairs, with a bottleneck feature vector of size 7 × 512 which exactly matches the dimensionality of the style vector used in mask-to-3D module. This design choice ensures that the network operates at the same representational capacity as the downstream 3D module. During training, we apply a Style Vector loss that encourages the bottleneck features to match the style vectors extracted from ground‐truth masks, to transfer the semantic priors of pretrained mask-to-3D module into sketch-to-mask module. As a result, our network bridges the domain gap between sketch inputs and segmentation outputs, enabling 3D model generation from a single sketch.



\subsection{Mask-to-3D Module}

We adopt the Seg2Face and Seg2Cat models from pix2pix3D~\cite{pix2pix3d} as our mask-to-3D modules for each task.
It is a 3D-aware conditional generative model conditioned on 2D label maps, such as segmentation or edge maps. By leveraging a tri-plane feature map, pix2pix3D generates volumetric representations of semantic labels and appearance, which can be rendered from arbitrary viewpoints.
Given a 2D label map \( I_s \), the framework first encodes the input using a conditional encoder \( E \), which maps the label map \( I_s \) and a random latent code \( z \sim \mathcal{N}(0, I) \) to a latent style vector \( w^+ \):
\begin{equation}
    w^+ = E(I_s, z).
\end{equation}

The style vector \( w^+ \) is then used to modulate a hybrid 3D representation parameterized by tri-planes and a multi-layer perceptron (MLP), enabling the generation of 3D outputs. 
The MLP computes color \( c \in \mathbb{R}^3 \), density \( \sigma \in \mathbb{R}^+ \), feature vectors \( \phi \in \mathbb{R}^l \), and semantic labels \( s \in \mathbb{R}^c \), where \( c \) is the number of classes.
To generate a 2D output from a novel viewpoint, pix2pix3D employs volumetric rendering:
\begin{equation}
    I_c(r) = \sum_{i=1}^N \tau_i c_i, \quad I_s(r) = \sum_{i=1}^N \tau_i s_i,
\end{equation}
where \( \tau_i \) represents the transmittance probability of a photon along the ray \( r \).
Subsequently, the rendered images and their corresponding feature vectors \(\phi\) are provided as input to a CNN-based up-sampling module, which produces high-resolution RGB images \( \hat{I}_c^+ \) and semantic maps \( \hat{I}_s^+ \).


\subsection{Losses}

To train the network, we jointly apply a Style Vector loss and a Segmentation Loss. The Style Vector loss enforces alignment between the U-Net's bottleneck representation and the latent style vector 
\(w^+\) of pix2pix3D, while the Segmentation Loss ensures accurate transformation of sketches into face segmentation masks.

Let \(w^E\) denote the bottleneck embedding, and let \(w^+\) be the style vector of pix2pix3D style encoder. We then define the Style Vector loss as the mean squared error (MSE) between these two vectors:
\vspace{-7pt}
\begin{equation}
    \mathcal{L}_{\text{SV}} = \| w^{+} - w^{E} \|_2^2,
\end{equation}
By minimizing \(\mathcal{L}_{\text{SV}}\), we force the sketch-to-mask module to embed sketches into the same style latent space as segmentation masks, thereby enhancing its ability to reconstruct masks faithfully
For Segmentation Loss, we employ the Cross-entropy loss and Dice loss \cite{Sudre_2017diceloss}.

\vspace{-10pt}

\begin{equation}
    \mathcal{L}_{\text{CE}}(y_{i,c}, \hat{y}_{i,c}) = - \frac{1}{n} \sum_{i=1}^{n} \sum_{c=1}^{C} y_{i,c} \log(\hat{y}_{i,c})
\end{equation}

\vspace{-15pt}

\begin{equation}
    \mathcal{L}_{\text{Dice}}(y_{i,c}, \hat{y}_{i,c})\!=\! 1 \!-\! \frac{1}{C} \!\sum_{c=1}^{C}\! \frac{2 \!\sum_{i=1}^{n}\! \hat{y}_{i,c} y_{i,c}}{\!\sum_{i=1}^{n}\! \hat{y}_{i,c}\!+\!\sum_{i=1}^{n}\! y_{i,c}\!+\!\epsilon}
\end{equation}
Where \( i \) denotes an individual pixel, \( n \) represents the total number of pixels, \( c \) refers to a specific segmentation class, and \( C \) is the total number of classes. \( y_{i,c} \in \{0,1\} \) indicates the ground truth label of the \( i \)-th pixel for class \( c \), while \( \hat{y}_{i,c} \) denotes the predicted probability for the same class. A small constant \( \epsilon \) is added to the denominator for numerical stability.

The overall training loss $\mathcal{L}_{\text{total}} = \mathcal{L}_{\text{SV}} + \mathcal{L}_{\text{CE}} + \mathcal{L}_{\text{Dice}}$ is computed by combining Style Vector loss, Cross-Entropy Loss, and Dice Loss.

\begin{table}[t!]
  \centering
  \begin{tabular}{@{} l l c c c @{}}
    \toprule
    Dataset & Method & FID $\downarrow$ & KID $\downarrow$ & FVV $\downarrow$ \\
    \midrule
    \multirow{2}{*}{CelebA} & pix2pix3D & 232.81 & 0.3142 & 0.20 \\
     & S3D (Ours) & \textbf{21.71} & \textbf{0.0065} & \textbf{0.18} \\
    \midrule
    \multirow{2}{*}{AFHQ} & pix2pix3D & 27.36 & 0.0054 & - \\
     & S3D (Ours) & \textbf{23.86} & \textbf{0.0047} & - \\
    \bottomrule
  \end{tabular}
  \caption{\textbf{Quantitative Results.} S3D demonstrated superior performance over the edge-to-3D model of pix2pix3D~\cite{pix2pix3d} for 3D generation of human and cat faces. The FVV metric was not measured for the AFHQ dataset, as it is specific to multi-view consistency of human faces.}
  \label{tab:quantitative}
  \vspace{-15pt}
\end{table}

\section{Experiments}

Our experimental results show that our model generates high-quality 3D models and achieves superior performance on quantitative metrics.
We used the Multi-Modal-CelebA-HQ dataset~\cite{liu2015deep, karras2017progressive, CelebAMask-HQ} for human face generation and the AFHQ~\cite{choi2020starganv2} dataset for training and testing in cat face generation.
Through ablation study, we validate our architectural decisions, with each component contributing to overall performance.
We employed the edge-to-3D model of pix2pix3D~\cite{pix2pix3d} as a baseline for both quantitative and qualitative evaluation.
The details of the augmentation strategy are provided in \autoref{section:augmentaion}.




\subsection{Qualitative Results}
\autoref{fig:teaser} presents a comparative analysis between our S3D approach and the baseline, highlighting differences in 3D generation quality. For human face generation, S3D successfully synthesizes both ground truth and novel viewpoints from input sketches, whereas pix2pix3D fails to converge on this task. In the case of cat faces, while pix2pix3D produces multi-view outputs, they exhibit lower fidelity and artifacts compared to the sharper, more coherent results achieved by S3D.
Additional qualitative results can be found in \autoref{section:moreresults}.

\subsection{Quantitative Results}
\autoref{tab:quantitative}
presents the quantitative results of S3D and pix2pix3D. For the evaluation metrics, FID~\cite{fid} and KID~\cite{kid} are employed to assess the fidelity of generated images. FVV Identity~\cite{pix2pix3d} is used to quantify multi-view consistency of the reconstructed human face. The results demonstrate that our S3D consistently outperforms the baseline across both datasets.

\begin{table}[t!]
  \centering
  \begin{tabular}{@{} l l c c @{}}
    \toprule
    Dataset & Method & mIoU $\uparrow$ & mAP $\uparrow$ \\
    \midrule
    \multirow{2}{*}{CelebA} 
     & w/o \(\mathcal{L_\text{SV}}\) & 0.692 & 0.793 \\
     & w/ \(\mathcal{L_\text{SV}}\)                  & \textbf{0.698} & \textbf{0.823} \\
    \midrule
    \multirow{2}{*}{AFHQ} 
     & w/o \(\mathcal{L_\text{SV}}\) & 0.804 & 0.884 \\
     & w/ \(\mathcal{L_\text{SV}}\)                 & \textbf{0.807} & \textbf{0.890}  \\
    \bottomrule
  \end{tabular}
  \caption{\textbf{Ablation Study Results.} Evaluation of mIoU and mAP for our S3D framework with and without the Style Vector loss $\mathcal{L}_{\text{SV}}$ on CelebA and AFHQ datasets.}
  \label{tab:ablation_miou_map}
\end{table}

\subsection{Ablation Study}

We compare our full model with an ablated variant without the Style Vector loss $\mathcal{L}_{\text{SV}}$. To quantify the contribution of this loss, we evaluate the quality of the output in terms of the performance of the sketch-to-mask module. Our full model achieves superior mask inference results, indicating that the Style Vector loss $\mathcal{L}_{\text{SV}}$ significantly improves the accuracy of semantic mask generation.

\section{Conclusion}

We present S3D, a one-shot 3D model generation pipeline. Our key contributions include a Style Vector loss that improves consistency between sketch-to-mask module and mask-to-3D module intermediate representations, complemented by segmentation losses for enhanced mask accuracy. Additionally, we found that our data augmentation strategy contributes significantly to the quality and robustness of the generated results.
This approach enables the previously unattainable task of generating 3D human faces directly from facial sketches.

Although S3D demonstrates strong performance in 3D face generation, our study reveals limitations. In particular, a performance gap emerges due to limited diversity in sketches with attributes such as long hair. To address this, we plan to expand the dataset and investigate debiasing techniques as part of our future work.

~\\
\textbf{Acknowledgements   }
This work was supported by Institute of Information \& communications Technology Planning \& Evaluation (IITP) grant funded by the Korea government(MSIT) (RS-2022-00143911,AI Excellence Global Innovative Leader Education Program). The researchers at Seoul National University were funded by the Korean Government through the grants from NRF (2021R1A2C3006659), IITP (RS-2021-II211343) and KOCCA (RS-2024-00398320).
{
    \small
    \bibliographystyle{ieeenat_fullname}
    \bibliography{main}
}

\clearpage
\setcounter{page}{1}
\maketitlesupplementary

\appendix
\renewcommand{\thesection}{\Alph{section}}
\renewcommand{\thesubsection}{\Alph{section}.\arabic{subsection}}

\section{Data Augmentation}
\label{section:augmentaion}


We designed a simple yet effective sketch-specific data augmentation strategy to improve generalization of the sketch-to-mask network. The core idea is to simulate structural perturbations in the sketches using morphological operations.

\subsection{Augmentation Details}

For each sketch input, we randomly apply one of three augmentation strategies: keeping the original sketch (with a probability of 50\%), applying dilation (25\%), or applying erosion (25\%). Dilation is performed using a kernel size of 3, while erosion uses a kernel size of 7. These augmentations are applied uniformly to both human and cat face sketches, introducing beneficial variability to improve robustness during training.

\subsection{Effects of Augmentation Strategy}

\vspace{-10pt}

\begin{figure}[htbp]
  \centering
  \includegraphics[width=\linewidth]{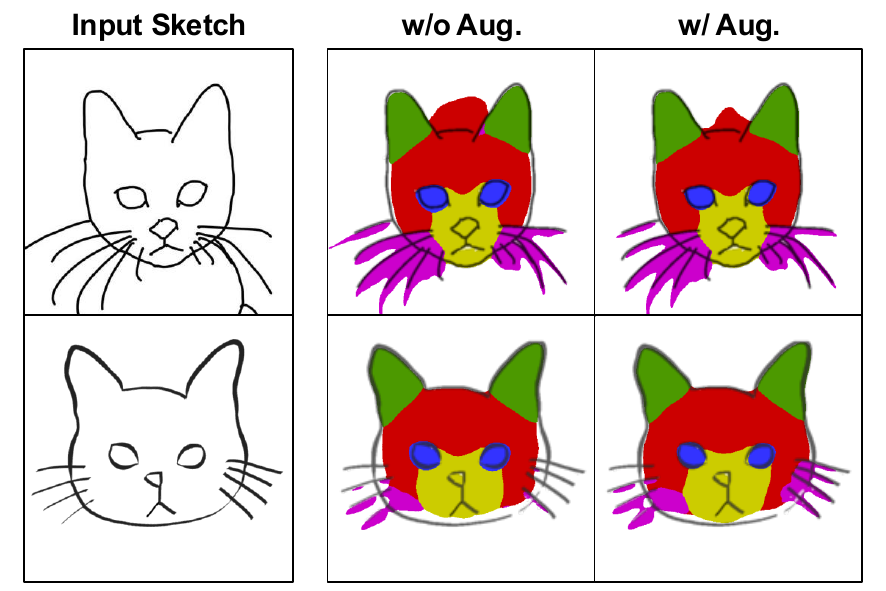}
  \caption{\textbf{Effect of data augmentation on training robustness.} Our sketch-specific augmentation strategy enables the generation of diverse sketch styles. Without data augmentation, the model often produces segmentation masks that are smaller than the true facial contours and fails to capture regions such as facial hair. After incorporating augmentation, it generates more robust masks from the input sketches.}

  \label{fig:aug_ablation_vis}
\end{figure}

\autoref{fig:aug_ablation_vis} illustrates the effect of our sketch-specific augmentation strategy by visualizing predicted segmentation masks before and after augmentation. Because agmented samples exhibit a wide range of stylistic variations, they encourage the network to generalize better and produce more robust mask predictions.

\autoref{tab:augmentation_ablation} illustrates how applying data augmentation affects the performance of the sketch-to-mask network. Although, on the CelebA dataset, the mean Intersection over Union (mIoU) was marginally higher without augmentation, overall augmentation led to performance improvements across both datasets.

\begin{table}[htbp]
  \centering
  \begin{tabular}{@{} l l c c @{}}
    \toprule
    Dataset & Method & mIoU $\uparrow$ & mAP $\uparrow$ \\
    \midrule
    \multirow{2}{*}{CelebA} 
     & w/o Augmentation & \textbf{0.699} & 0.818 \\
     & w/ Augmentation  & 0.698 & \textbf{0.823} \\
    \midrule
    \multirow{2}{*}{AFHQ} 
     & w/o Augmentation & 0.804 & 0.881 \\
     & w/ Augmentation  & \textbf{0.807} & \textbf{0.889} \\
    \bottomrule
  \end{tabular}
  \caption{\textbf{Effect of Data Augmentation.} Comparison of mIoU and mAP for models trained with and without sketch-specific augmentation on CelebA and AFHQ. Data augmentation yields improvements in sketch-to-mask performance.}
  \label{tab:augmentation_ablation}
\end{table}

\vspace{-15pt}
\section{Embedding Space Analysis}
\vspace{-10pt}

\begin{figure}[htbp]
  \centering
  \includegraphics[width=1\linewidth]{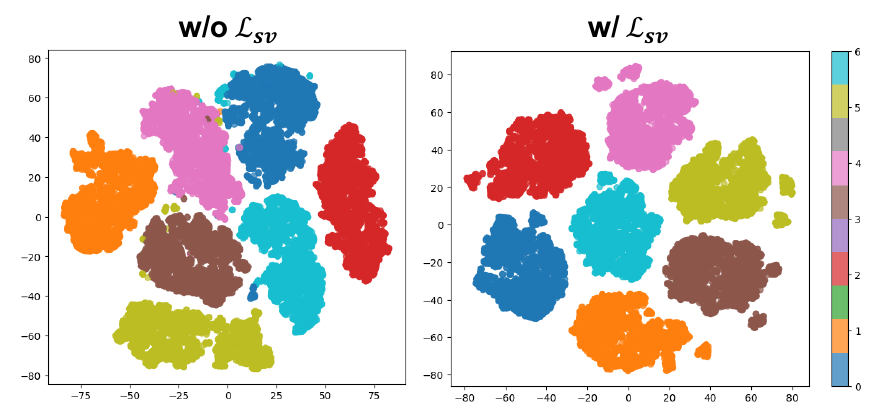}
  \caption{\textbf{t-SNE visualization of style embeddings.} Embeddings learned with the style vector loss $\mathcal{L}_{\text{SV}}$ (right) form tight and well-separated clusters, reflecting more structured representation learning. Without $\mathcal{L}_{\text{SV}}$ (left), the distributions are scattered and less semantically aligned.}
  \label{fig:tsne}
\end{figure}
To further investigate the impact of the style vector loss $\mathcal{L}_{\text{SV}}$, we visualize the learned embedding space using t-SNE in \autoref{fig:tsne}. The embeddings produced with $\mathcal{L}_{\text{SV}}$ form tighter and more coherent clusters, indicating that the style vector loss helps guide the network to organize semantic features in a more discriminative manner. Without this constraint, the embeddings are more scattered and inconsistent across samples, particularly within ambiguous sketch inputs.

\section{More Qualitative Results}
\label{section:moreresults}

\autoref{fig:more_qulitative_results} presents additional qualitative results for our proposed S3D model and the baseline. These results underscore the effectiveness of S3D in generating high-fidelity 3D models from sketches on both the CelebA and AFHQ datasets.


\twocolumn[{
    \maketitle
    \begin{center}
        \includegraphics[width=1\textwidth]{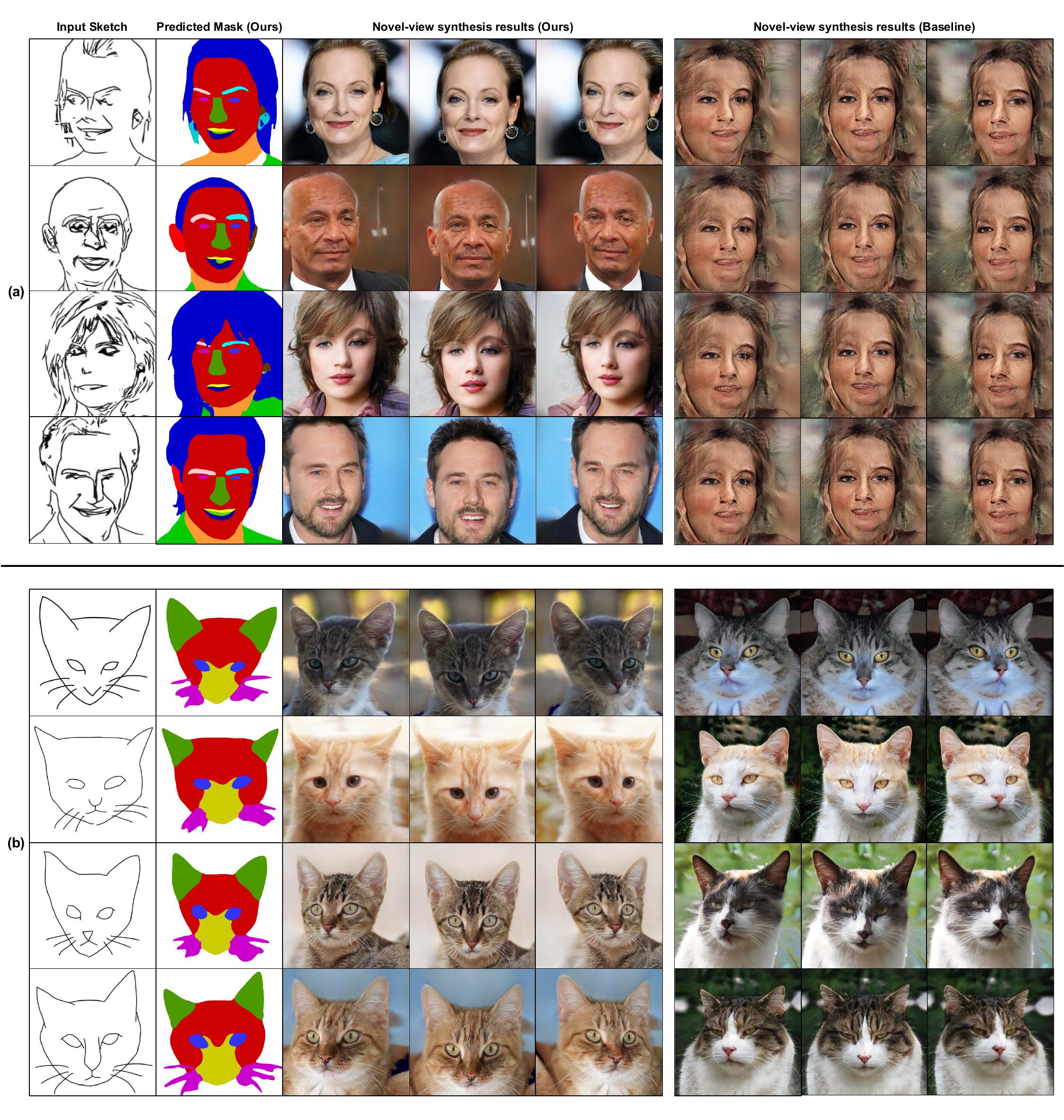}
        \captionof{figure}{
        \textbf{Additional qualitative comparisons between our S3D method and the baseline.} Across diverse sketch inputs, S3D consistently generates realistic, geometrically coherent 3D models, whereas the baseline produces unnatural structures misaligned with the sketches. (a) On sketch-to-human-face inputs, S3D recovers plausible 3D face geometry, while the baseline fails to form a coherent face. (b) On sketch-to-cat inputs, S3D generates high-quality 3D cat models aligned with the sketches, whereas the baseline yields distorted, misaligned shapes.
        }
        \label{fig:more_qulitative_results}
    \end{center}
}]

\end{document}